%% file: 00-main.tex
\documentclass[10pt,conference]{IEEEtran}
\usepackage{amsmath,amssymb,amsfonts}
\usepackage{algorithmic}
\usepackage{graphicx}
\usepackage{textcomp}
\usepackage{xcolor}
\usepackage{hyperref} 
\usepackage{todonotes}

\usepackage{adjustbox}
\usepackage{rotating}
\usepackage{multirow}
\usepackage{dblfloatfix}
\usepackage[
backend=biber,
style=ieee,
maxnames=5,
minnames=1
]{biblatex}
\def\cca#1{\cellcolor{black!#10}\ifnum #1>5\color{white}\fi{#1}}

\def\BibTeX{{\rm B\kern-.05em{\sc i\kern-.025em b}\kern-.08em
    T\kern-.1667em\lower.7ex\hbox{E}\kern-.125emX}}
\begin{document}

\title{A Comparative Analysis of Adversarial Robustness for Quantum and Classical Machine Learning Models}

\author{\IEEEauthorblockN{Maximilian Wendlinger\IEEEauthorrefmark{1}\IEEEauthorrefmark{2}, Kilian Tscharke\IEEEauthorrefmark{1}, Pascal Debus\IEEEauthorrefmark{1}}
\IEEEauthorblockA{\IEEEauthorrefmark{1}\textit{Quantum Security Technologies} \\
\textit{Fraunhofer Institute for Applied and Integrated Security}\\
Garching near Munich, Germany}
\IEEEauthorblockA{\IEEEauthorrefmark{2}\textit{Technical University of Munich}\\
maxi.wendlinger@tum.de, \{kilian.tscharke, pascal.debus\}@aisec.fraunhofer.de}
}

\maketitle

\begin{abstract}
Quantum machine learning (QML) continues to be an area of tremendous interest from research and industry. While QML models have been shown to be vulnerable to adversarial attacks much in the same manner as classical machine learning models, it is still largely unknown how to compare adversarial attacks on quantum versus classical models. In this paper, we show how to systematically investigate the similarities and differences in adversarial robustness of classical and quantum models using transfer attacks, perturbation patterns and Lipschitz bounds. More specifically, we focus on classification tasks on a handcrafted dataset that allows quantitative analysis for feature attribution. This enables us to get insight, both theoretically and experimentally, on the robustness of classification networks. We start by comparing typical QML model architectures such as amplitude and re-upload encoding circuits with variational parameters to a classical ConvNet architecture. Next, we introduce a classical approximation of QML circuits (originally obtained with Random Fourier Features sampling but adapted in this work to fit a trainable encoding) and evaluate this model, denoted Fourier network, in comparison to other architectures. Our findings show that this Fourier network can be seen as a ``middle ground'' on the quantum-classical boundary. While adversarial attacks successfully transfer across this boundary in both directions, we also show that regularization helps quantum networks to be more robust, which has direct impact on Lipschitz bounds and transfer attacks.
\end{abstract}


\input{01-intro}
\input{03-related-work}
\input{02-theoretical-background}
\input{04-experimental-setup}
\input{06-summary-outlook}
\input{07-code}
\printbibliography
\clearpage

\input{99-appendix}

\end{document}

%% file: 01-intro.tex
\section{Introduction}

Quantum machine learning (QML) offers the potential to expand the current frontiers of many fields in computational technologies \cite{sarma2019machine}, with its power to leverage quantum-mechanical effects such as entanglement and high-dimensional latent spaces paired with insights from classical machine learning (ML). As a result, recent years have seen an immense research interest in different possibilities to leverage these effects for practical challenges. Most prominently, this includes parameterized quantum circuit (PQC) architectures that allow for the training of variational parameters inside the quantum circuit in order to fit a specific function for classification or regression tasks. It has been shown that although results from quantum variational classifiers look promising, the models can be fooled with carefully crafted modifications of the input samples \cite{lu2020quantum}. These adversarial perturbations, allowing an attacker to force a high misprediction rate in the attacked model, are investigated in the emerging field called quantum adversarial machine learning (QAML). For classical machine learning approaches, adversarial attacks have been rigorously explored, opening the door for research to find similarities and differences between adversarial attacks on classical and quantum machine learning architectures.

In an effort to make the first steps in this direction, West et al. \cite{west2023benchmarking} investigate the behavior of different classical and quantum ML models under adversarial attacks and the resulting perturbation patterns in the input. Interestingly, the authors found that classical attacks fail to transfer to the PQC architecture while attacks originally designed for the quantum model seem to work for classical ML models. In this light, the authors suspected a "quantum supremacy" in adversarial robustness. In this work, we aim to continue this line of research by connecting theoretical and practical insights from previous work with experiments on a specifically designed dataset. The contributions of this paper are as follows:

\begin{itemize}
    \item We create a four-class image dataset for QML that allows semantic analysis while keeping input dimensions low
    \item On this dataset, attack patterns resulting from adversarial attacks on PQC models are evaluated
    \item Transfer attacks between quantum models of different architectures are conducted
    \item We construct a classical Fourier network similar to \cite{schreiber2023classical} and a convolutional network to compare attack patterns and check quantum/classical transferability
    \item Regularization is used to increase the robustness of quantum models and verify the impact on attacks
    \item We link outcomes of the experiments with theoretical Lipschitz bounds for classical and quantum models
\end{itemize}
We start by presenting relevant work related to the topics discussed here before briefly introducing the theoretical concepts used in the paper. In Section \ref{sec-4-exp}, we describe the experimental setup and results, followed by a short summary of the findings.

%% file: 03-related-work.tex
\section{Related work} \label{sec-3-rel-work}
Quantum adversarial machine learning has been investigated thoroughly, both in theory (e.g. \cite{lu2020quantum, gong2022enhancing, }) and practice (e.g. \cite{ren2022experimental, west2023benchmarking}, also see \cite{west2023towards} for a review on current progress and challenges in QAML). An important aspect of quantum machine learning, especially in QAML, is to find parallels between insights from classical ML models (in this case the vulnerability to adversarial attacks) and observations from experiments from QML. 

As a first step, West et al. \cite{west2023benchmarking} benchmarked classical Convolutional networks such as ResNet \cite{he2016deep} against quantum variational circuit architectures of different sizes. Specifically, the authors used the transferability of the attacks across the quantum/classical boundary and the visual appearance of the attacks (discussed in Section \ref{sec-4-exp}) in images of different datasets (FashionMNIST, CIFAR10, and CELEB-A) to conclude that quantum models may offer supremacy in robustness "by design". However, the training of the quantum models itself and the effect of hyperparameters (training time, learning rate, regularization) are not taken into account. Also, the data is encoded into the models using amplitude encoding, which is not yet feasible to implement on real quantum hardware for large input dimensions. In this work, we will therefore fill in these gaps by systematically investigating quantum models of different architectures (re-upload encoding and amplitude encoding) and their relation to classical adversarial machine learning with respect to training time and regularization.

Bridging the gap between quantum models and classical ML techniques is also one of the main themes of classical approximation papers \cite{landman2022classically, schreiber2023classical, sweke2023potential} that use the function classes of PQC models as a basis to construct a quantum kernel and classically optimize over embedded features, as discussed in Section \ref{sec-theo}. Here, we use insights from this corpus of work to construct a classical approximation of our PQC model, that is used as a ``middle ground'' between purely classical models such as ConvNet or ResNet architectures and quantum models such as variational circuits. We see that we can indeed make statements about the robustness of the quantum model by analyzing the classical counterpart, which is often easier to examine.

These investigations of generalization and adversarial robustness are done both theoretically via Lipschitz bounds and practically via transfer attacks and attack pattern analysis on a constructed image dataset. Lipschitz bounds (as detailed in Subsection \ref{subsec-advattacks-lip}) have been motivated and derived for classical machine learning \cite{virmaux2018lipschitz, fazlyab2019efficient, pauli2021training} -- although only approximated Lipschitz bounds can be calculated for typical neural network architectures -- and for quantum machine learning \cite{berberich2024robustness, berberich2023training}. In the QML domain, results on Lipschitz bounds have only been tested for simple, low-dimensional datasets, therefore it will also be a focus of this paper to incorporate the calculation of Lipschitz bounds of classical and quantum models for real-world multi-class classification tasks into the analysis of robustness and generalization capability.

%% file: 02-theoretical-background.tex
\section{Theoretical background} \label{sec-theo}

\subsection{Function classes of parametrized quantum circuits} \label{subsec-funcclasses}
Parametrized quantum circuits typically consist of layers of rotational and entangling gates, followed by measurements of the first $k$ qubits. Classical input features are generally embedded into the latent model space of noisy intermediate-scale quantum (NISQ) devices using some form of repeated rotational encoding \cite{PerezSalinas2020datareuploading}. A notable second option -- although not yet feasible in practice due to the large depth of the encoding circuit -- is to encode the (possibly padded) $2^{n}$ input features into the amplitudes of the n-qubit quantum state via amplitude embedding. Both of these variants will be explored in Section \ref{sec-4-exp}. In the following, we consider a typical re-upload encoding circuit \cite{PerezSalinas2020datareuploading, schuld2021effect}, where the rotational parameters $\theta$ of the PQC are optimized in a training loop to minimize some suitable loss function. As model output, we consider the expectation value of the measured qubits, resulting in the expression
\begin{equation} \label{eq-model-fct}
f(x ; \theta)=\left\langle 0\left|U(x;\theta)^{\dagger} \mathcal{M} U(x;\theta)\right| 0\right\rangle
\end{equation}
for some data input vector $x = (x_1,\ldots,x_D)$, a parametrized circuit represented by some unitary $U(x;\theta)$ and an observable $\mathcal{M}$.
We can write the functions represented by such PQC models as a truncated Fourier series of the form
\begin{equation} \label{eq-truncfourier}
f(x ; \theta)=\sum_{\omega \in \Omega} c_\omega e^{i \omega x}
\end{equation}
where the accessible frequencies $\omega \in \Omega$ are determined by the data encoding strategy and the Fourier coefficients $c_\omega$ result from the trainable layer blocks \cite{schuld2021effect, vidal2020input}. In Section \ref{sec-4-exp}, we construct a trainable encoding that increases the accessible frequency spectrum \cite{PerezSalinas2020datareuploading, Shin2023exponential} and show the implications for adversarial robustness.

\subsection{Classical approximations of PQC models} \label{subsec-rff}

In the light of Subsection \ref{subsec-funcclasses}, it seems natural to try and find classical approximates for the function class that a PQC can realize.
Indeed, we can regroup the terms in the partial Fourier series as done in \cite{landman2022classically, sweke2023potential} by defining 
$$
\begin{gathered}
a_\omega:=c_\omega+c_{-\omega} \in \mathbb{R} \\
b_\omega:=\frac{1}{i}\left(c_\omega-c_{-\omega}\right) \in \mathbb{R}
\end{gathered}
$$
to write Equation \eqref{eq-truncfourier} as
\begin{equation} \label{eq-absum}
\begin{aligned}
f(x ; \theta) & =\sum_{\omega \in \Omega_{+}} c_\omega e^{i \omega x}+c_{-\omega} e^{-i \omega x} \\
& =\sum_{\omega \in \Omega_{+}} a_\omega \cos (\omega x)+b_\omega \sin (\omega x)
\end{aligned}
\end{equation}
where $\Omega_{+}$ contains the positive half of the frequencies in $\Omega$ \cite{landman2022classically}. If the frequency spectrum is known and small enough to allow efficient computation of all summands, we can construct a classical surrogate model \cite{schreiber2023classical}. In most cases, however, the frequency spectrum is not accessible in its entirety due to the exponential number of elements (e.g., considering Pauli encoding gates, $\left|\Omega_{+}\right|\in\mathcal{O}(L^{D})$ for $D$ input dimensions and $L$ re-upload encoding layers \cite{caro2021encoding,landman2022classically}). To assemble possible frequencies $\omega \in \Omega_{+}$ for classical approximations of PQC architectures, recent work utilizes Random Fourier features (RFF) sampling \cite{landman2022classically, sweke2023potential}. The values of the classical parameters $a_\omega$ and $b_\omega$ are then found by classical optimization techniques (in \cite{landman2022classically}, this corresponds to a ridge regression optimization). In this work, we use a single-layer feed-forward network with the corresponding sine and cosine functions inside the hidden layer, similar to the approach described in \cite{schreiber2023classical}. For PQC models defined as in \eqref{eq-truncfourier}, one can then simply RFF-sample frequencies $\omega \in \mathbb{R}^{D}$ according to \cite{landman2022classically}, construct the (classical) sum from Equation \eqref{eq-absum} and use stochastic gradient descent to find a set of optimal parameters. For trainable encoding architectures discussed in Section \ref{sec-4-exp}, we optimize over the encoding weights as well as the $a_\omega$ and $b_\omega$ parameters. The precise model architectures (for the classical and quantum models), as well as the training and optimization specifications can be found in Appendix \ref{app-architectures}.

\subsection{Adversarial attacks and Lipschitz bounds} \label{subsec-advattacks-lip}

\subsubsection{Adversarial attacks on QML}

Quantum machine learning models are generally trained using hybrid quantum-classical approaches where the PQC is evaluated on (actual or simulated) quantum hardware in contrast to the classical optimization algorithm of evaluating loss functions and updating the model parameters. This allows us to define attacks on quantum models using the same principles that are known in classical machine learning. As such, QAML finds perturbations $\delta\in\Delta$ to individual input samples $x =(x_1, \ldots x_D)$ maximizing the loss function $\mathcal{L}$ as
\begin{equation}
\delta \equiv \; \underset{\delta^{\prime} \in \Delta}{\operatorname{argmax}} \;\mathcal{L}\left(f\left(x+\delta^{\prime} ; \theta^*\right), y\right)
\end{equation}
where $f: \mathbb{R}^{D} \to \mathbb{R}^K$ is the PQC model mapping input samples $x$ to softmax probability distributions of $K$ classes (with ground-truth labels $y$) via measurements of the first $K$ qubits. For an adversarial attack to be considered useful, it must hold that the modifications to the input elements are imperceptible, i.e. that $\Delta=\{\delta \in \mathbb{R}^{D}: \| \delta\|_{\infty} \le \varepsilon\}$\footnote{It is common practice to use the $\ell_{\infty}$ norm for adversarial attacks, however other norms are also studied in literature.}, where $\varepsilon$ denotes the perturbation strength of the attack. Using these perturbations $\delta$ on a different model than that of the attack is known as conducting a \textit{transfer attack}. It has been shown that many classical models are vulnerable to attacks even from source models that fundamentally differ in architecture \cite{papernot2017practical, demontis2019adversarial, ilyas2019adversarial}.

\subsubsection{Lipschitz bounds for robustness estimation} 
A general strategy to evaluate the robustness of (classical and quantum) models is to investigate the effect of input differences on the model output via Lipschitz bounds. The Lipschitz bound of a function $f$ is commonly defined as the smallest constant $L$ satisfying
\begin{equation}
\left\Vert f(x)-f(y)\right\Vert\le L \left\Vert x-y\right\Vert    
\end{equation}
where one can directly derive a bound on the worst-case effect of an adversarial perturbation by setting $y=x+\delta$.

While an exact computation of global Lipschitz bounds for classical ML models is NP-hard \cite{virmaux2018lipschitz, fazlyab2019efficient}, different approaches exist to approximate $L$. One of the most prominent ideas makes use of semi-definite programming methods to solve an inequality arising from the slope-restriction of the activation functions used in the network \cite{fazlyab2019efficient, pauli2021training}. More precisely, this means that the slope-restriction of an activation function $\varphi:\mathbb{R}\to\mathbb{R}$ given by
\begin{equation}
\alpha \le \frac{\varphi(y)-\varphi(x)}{y-x} \le \beta \quad \forall x,y\in \mathbb{R}  
\end{equation}
can be reformulated into an incremental quadratic constraint for $\varphi$, which gives rise to Theorems 1 and 2 in Fazlyab et al. \cite{fazlyab2019efficient} stating that one can find efficient and accurate approximations for the Lipschitz bounds of a model if we have access to all weights and activation functions.

For QML models, recent work by Berberich et al. \cite{berberich2023training} has shown that the linearity of quantum models in feature space (defined e.g. by the space of complex matrices and their Hilbert-Schmidt inner products in Schuld \cite{schuld2021supervised}) can be used to compute tight Lipschitz bounds, allowing systematic evaluation and robust training of quantum circuits. Concretely, the authors derive the bound
\begin{equation}
L_{\Theta}=2\left\Vert\mathcal{M}\right\Vert\sum_{j=1}^{N}\left\Vert w_j\right\Vert \left\Vert H_j\right\Vert    
\label{eq-lip-quant}
\end{equation}

where $\mathcal{M}$ is the measurement observable, $H_j$ can be any encoding Hamiltonian (for the experiments, Pauli gates are applied) and the weights $w_j$ are used in their trainable encoding gates of the form $U_{j,\Theta_j}(x)=e^{-i(w_j^{\top}x+\theta_j)H_j}$.  

This notion of Lipschitz bounds for PQC models using trainable encodings can directly be used to construct a regularized loss function, where a trade-off between the loss target and the norm of encoding weights and Hamiltonians emerges. In the experiments, we make use of the Lipschitz regularizer presented in \cite{berberich2023training} for enhancing the robustness of our quantum models. The optimization problem of the PQC model in \cite{berberich2023training} is then described by
\begin{equation}
\min _{\Theta} \frac{1}{n} \sum_{k=1}^n \mathcal{L}\left(f\left(x_k;\theta\right), y_k\right)+\lambda \sum_{j=1}^N\left\|w_j\right\|^2\left\|H_j\right\|^2 .    
\end{equation}

In Section \ref{sec-4-exp}, we use Lipschitz bounds for comparison between classical and quantum ML models, where we see that the Lipschitz bounds correlate with the ability to transfer attacks. Specific derivations of Lipschitz bounds for the PQC architectures we examine can be found in Appendix \ref{app-lip}.

%% file: 04-experimental-setup.tex
\section{Experimental setup and Results} \label{sec-4-exp}
In this chapter, a brief overview of the dataset and models used in the experiments is given before constructing attacks on each model and checking the adversarial robustness, transferability, and feature attribution. To give a theoretical background for the results of the experiments, Lipschitz values and regularization of quantum models are introduced. For all experiments, an image classification task is conducted, where the output of the respective model corresponds to a probability distribution over the four possible classes. A cross-entropy loss function is used to optimize over parameters, evaluation is done via cost and accuracy scores. Classical models (RFF approximation model and ConvNet) are implemented using Pytorch \cite{paszke2019pytorch}, and quantum models are built with the Pennylane \cite{bergholm2018pennylane} library. This dataset contains four classes of grayscale images, where each pixel value lies in the interval $[0,1]$, analogous to the MNIST set \cite{lecun2010mnist}.
\subsection{Dataset and Preprocessing} \label{subsec-dataset}
To have some quantifiable measures to compare the shapes emerging from adversarial attacks on the different model architectures, we construct an artificial image dataset for classification. 
\begin{figure}[htbp]
\centerline{\includegraphics[width=0.5\textwidth]{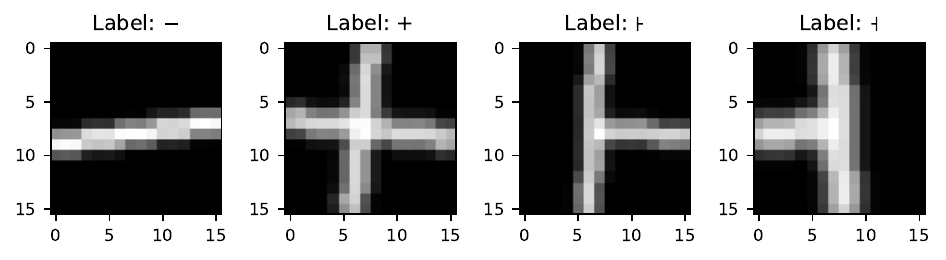}}
\caption{Examples of images in the corresponding classes of the synthetic dataset used for image classification.}
\label{fig-dataset}
\end{figure}

One sample of each class is shown in Fig. \ref{fig-dataset}, where the shape depicted in an image corresponds to the symbol in the respective label class (class 0: `$-$', class 1: `$+$', class 2: `$\vdash$', class 3: `$\dashv$'). This allows us to define the semantic meaning of pixel perturbations on the image and distinguish attacks that visually alter the correct label from perturbations that look like random noise. Each pair of classes has some defined attack perturbation -- to flip the classification output from one to the other class -- which humans would consider meaningful (e.g. adding a vertical bar to visually change a `$-$' into a `$+$'). In the work conducted by West et al. \cite{west2023benchmarking}, these semantically meaningful perturbations (on their datasets such as FashionMNIST) were found predominantly on attacks on quantum models, therefore the authors suspected that quantum models have the capability to attend to more sensible, robust features in the input data. In the following subsections, we investigate that claim on the constructed dataset that offers more insight into the semantic meaning of attacks.

The training dataset (test dataset) consists of 1000 (200) images that are balanced in the classes and randomly permuted before training. Random rotations of $\pm10$ degrees, blurs, and different thicknesses in the bars help diversify the sets. The dataset is normalized if needed for the repective model (amplitude embedding) and reshaped into a vector if suitable for the model architecture. The different architectures are described in the following.

\subsection{Models in comparison}

\subsubsection{Parametrized Quantum Circuits}
The first model we train is a re-upload encoding variational classifier, where each input $x = (x_1,\ldots,x_D)$ is split into 3-tuples to facilitate integration into the angles of rotational gates. More precisely, for an input image of $(16\times 16)=256$ dimensions, we use 8 qubits and 32 layers, each consisting of a block of rotational gates followed by an entangling layer, giving $(32\cdot 8\cdot 3)=768$ variational parameters and a three times re-upload encoding. Each input feature is multiplied by some encoding weight, which we optimize alongside the variational parameters, adding up to a total of 1536 trainable parameters in the model. For the model layers, the Pennylane template \textit{StronglyEntanglingLayers}\footnote{\url{https://docs.pennylane.ai/en/stable/code/api/pennylane.StronglyEntanglingLayers.html}} is used.

As an alternative, we implement a variational model analogous to the previous one, with the difference that the input samples are encoded with Pennylane's \textit{AmplitudeEmbedding}\footnote{\url{https://docs.pennylane.ai/en/stable/code/api/pennylane.AmplitudeEmbedding.html}} routine. As a result, the rotational layers after the embedding subcircuit only contain the variational parameters without any input dependency or input weighing. Both models are visualized and described in more detail in Appendix \ref{app-architectures}. 

\subsubsection{ConvNet}
As a classical counterpart that has been used for decades in the realm of image classification, we construct a simple convolutional network (ConvNet) consisting of a single convolutional layer (including MaxPool operation and ReLU activation) followed by a linear layer with four output nodes. The simplicity of the network allows us in the following chapters to efficiently compute Lipschitz bounds using linearization techniques discussed in Section \ref{subsec-advattacks-lip} and Appendix \ref{app-lip-class}.

\subsubsection{Fourier network}
In addition to the models above, we train and evaluate a second classical model that is mathematically similar to the re-upload encoding quantum model. As introduced in Subsection \ref{subsec-rff}, we can approximate the parametrized circuit by a single-layer feedforward neural network learning a classical Fourier series 
\begin{equation}
g(x;w)=\sum_{w \in W} a_w \cos (w x)+b_w\sin (w x)
\label{eq-class-fourier}
\end{equation}

where the weights $w\in W$ (whose shape is determined by the number of hidden neurons in the inner layer) control the number of summands in $g(x;w)$. If we do not restrict ourselves to sampling $M$ frequencies $\omega$ as in \cite{landman2022classically} and setting $w_i=\omega_i$, but instead let the classical model also optimize over the encoding weights, we get a similar notion to the trainable encoding of the PQC architecture, albeit with a limited number of frequencies to learn. We can then optimize over both $w$ and $a_w,b_w$ (see Fig. \ref{fig-fourier-net} in Appendix \ref{app-architectures} for more information). A similar model architecture has been investigated as early as 1988 by Gallant and White \cite{gallant1988there}. We adhere to their naming and subsequently refer to our second classical network as the ``Fourier network''. As discussed before, this network structure can be seen as a ``middle ground'' between the mathematical description of a PQC architecture and a classical feedforward network such as ConvNet. 

\subsection{Model training and adversarial attacks}

First, all models (ConvNet, Fourier net, and both PQC architectures) are trained on the training dataset consisting of 1000 images as described in Subsection \ref{subsec-dataset}. In all cases, an AdamOptimizer routine is used with a learning rate of 0.001, no weight decay. Every model achieved a near-perfect accuracy on the (unsurprisingly easy) task of 4-class image classification. For most parts of the following chapters, we evaluate and attack each model after 20 epochs of training, however, we also conduct specific experiments concerning training epochs, therefore we also evaluate some models after training for 100 epochs. Whenever we use models trained for this increased duration, we explicitly mention it in the text. For visual confirmation of the models' ability to achieve correct classification results, the reader is referred to Appendix \ref{app-architectures}, where we show the performance plots for the models.

\begin{figure}[htb] 
\centering
\includegraphics[width=0.48\textwidth]{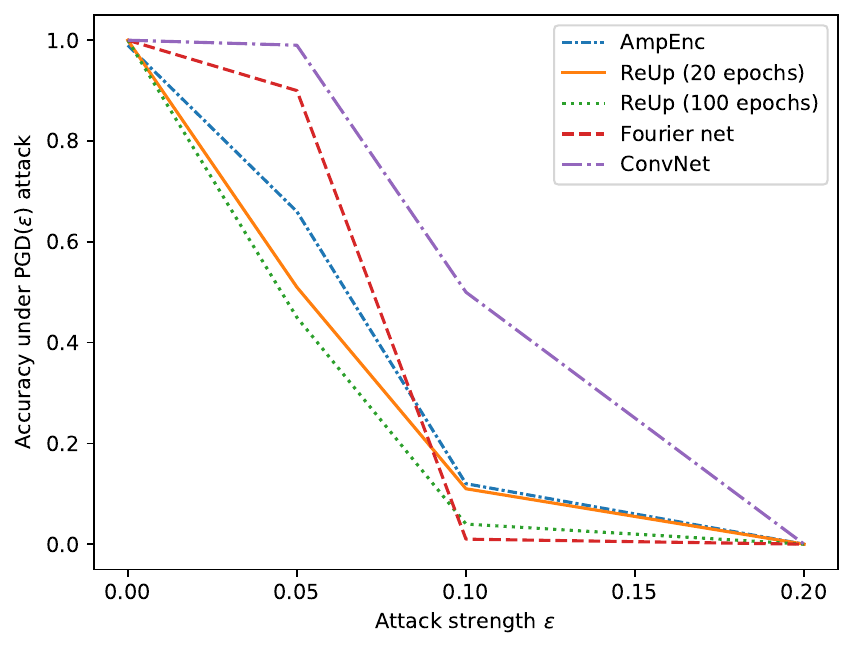}
\caption{Accuracy under PGD($\varepsilon$) attack for each investigated model. While all models are susceptible to attacks, the number of training epochs of the re-upload encoding model makes a difference in robustness.}
\label{fig-pgd}
\end{figure}

Next, we construct projected gradient descent (PGD) attacks \cite{madry2017towards} for each model, with varying attack strengths $\varepsilon\in \{0.05, 0.1, 0.2\}$. For each attack, 100 images from the dataset are modified according to the defined attack and fed into the model. The resulting accuracy under attack (i.e. accuracy with respect to perturbed input) is shown in Fig. \ref{fig-pgd}.
As expected, the performance of all models drastically deteriorates for relatively small changes in the input. Interestingly, we can see a difference in adversarial performance between the re-upload encoding model trained for 20 epochs and the one trained for 100 epochs, where the robustness of the model in question decreases during training. The changes in adversarial performance with increasing training time can also be verified when we investigate the different outcomes of transfer attacks in the following subsections.

\begin{figure}[htb] 
\includegraphics[width=0.49\textwidth]{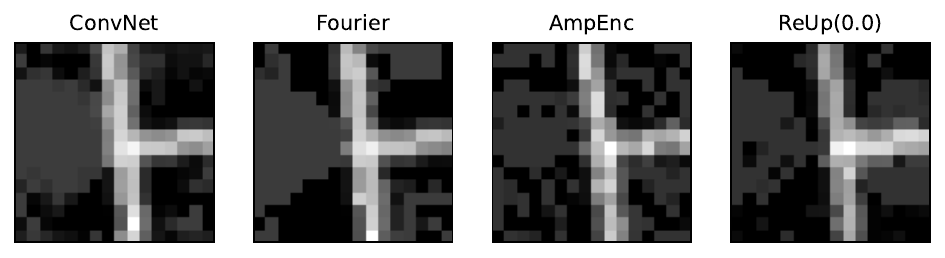}
\caption{Perturbed images resulting from PGD($0.1$) attacks for each model. The attacks clearly add a bar ($\pm 10$ degrees) to the left of each image, however differences in areas that do not bear semantic meaning are visible, e.g. in the amplitude encoding model we observe perturbations scattered across all parts of the image.}
\label{fig-imgpluspert}
\end{figure}

An adversarial sample including attack perturbations $x^{adv}=x+\delta$ is shown for each model in Fig. \ref{fig-imgpluspert}, where for ease of notation we drop the indices when referring to a single input sample and denote the components of $\delta$ by $\delta_i$. The resulting $\delta$ are obtained by PGD(0.1) attacks, i.e. the modifications shown in the images are in the range $[-0.1, 0.1]$.

All models misclassify the respective image as being in class `$+$' instead of the true class `$\vdash$'. Here we can see the reason for the chosen dataset: If the target of the attack is to fool the classifier into classifying a label `$+$', the most sensible modification would be to introduce a horizontal half-bar left of the vertical middle bar. Indeed, for each model, we find this change (in the range of the random $\pm 10\deg$ rotations) on the left part of the respective image. However, we can also spot differences in the level of ``random noise'' to the right of the middle bar in each image, in regions where a robust model should not be influenceable. For the classical networks, the attack is more locally bounded to the actual region of the label flip, whereas the quantum models' attack perturbations are noisier. Most strikingly, this can be verified in the amplitude encoding architecture, where the attack modifications are scattered across the whole image, and the overlap of the resulting attack pattern and the one a human would consider sensible is small. When introducing regularization of the re-upload encoding PQC models below, we return to the attack pattern analysis and show that regularization helps the model in finding sensible features in the input data.

\subsection{Transfer attacks}
\begin{figure}[htb]

\centering

\includegraphics[width=0.49\textwidth]{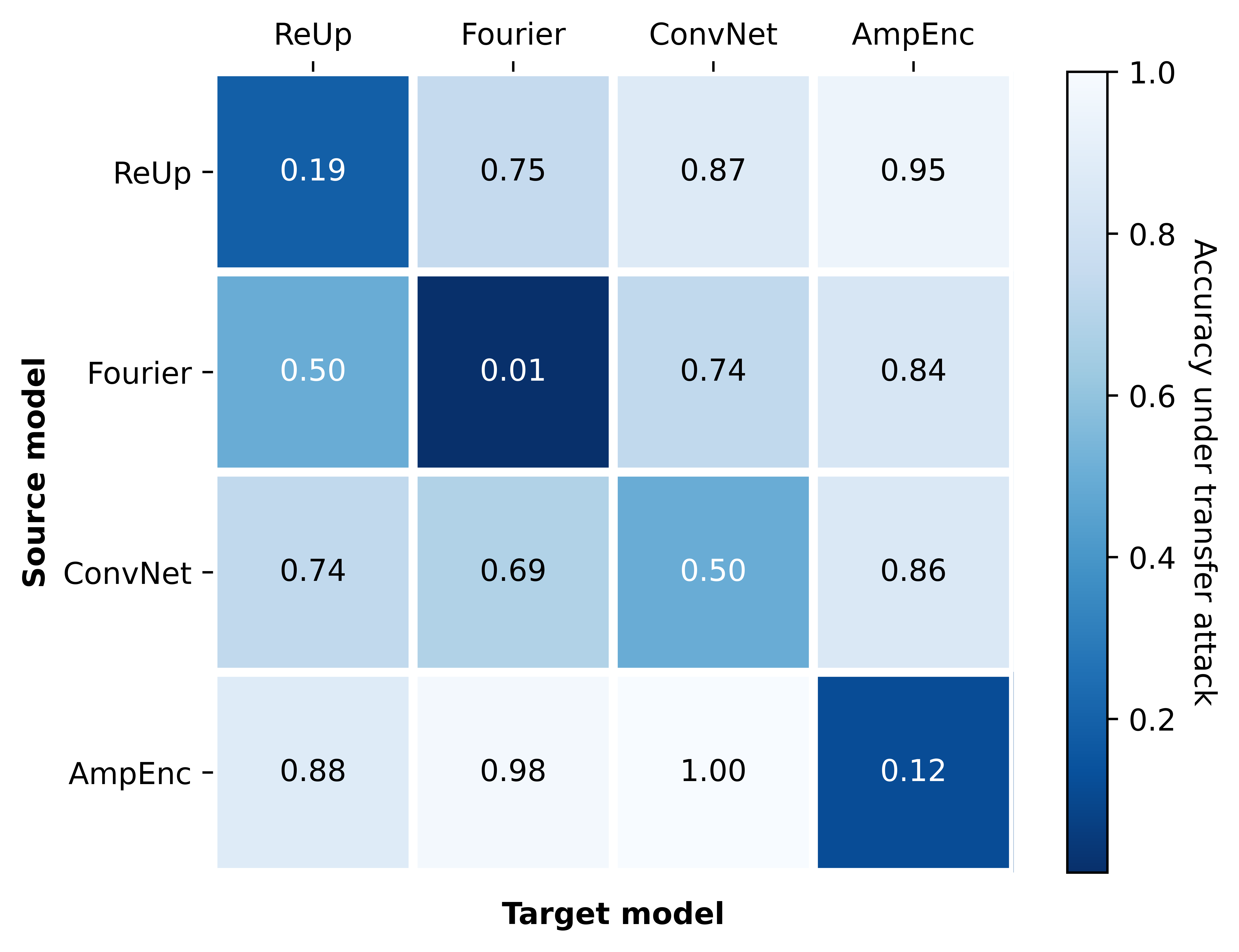}
\caption{Accuracies under PGD($0.1$) transfer attacks from source model (row) to target model (column) for each pair of models in question. Attacks tranfer notably better between ConvNet, Fourier net and re-upload encoding models than from/to amplitude encoding models.}
\label{table-transfatt}
\end{figure}

As a baseline for further investigations, we conduct experiments on the transferability of adversarial attacks. More specifically, we use the PGD(0.1) attacks (100 perturbed input images) as established above for each epoch 20 model and feed them into each of the other models (also trained for 20 epochs). The resulting accuracies are listed in Fig. \ref{table-transfatt}, where $entry[i,j]$ corresponds to the accuracy of model $j$ for transfer attacks from model $i$. 

The first observation can be made on the diagonal, where the vulnerability of each model to attacks constructed for the particular model itself is shown (corresponding to the values in Fig. \ref{fig-pgd} for $\varepsilon=0.1$). It is not surprising to see the effectiveness of the attacks for each model, however, it may be noted that the adversarial accuracy of the ConvNet stays relatively high (which can be due to the low parameter count in the convolutional architecture). More importantly, we can see that we can indeed construct adversarial attacks on classical networks that transfer well into the quantum domain, where a transfer attack of the Fourier network achieves a degradation of the re-upload encoding architecture by $0.5$. Also, we notice that although the amplitude encoding model seems robust to transfer attacks from the classical and quantum domain (last column), attacks constructed for the amplitude encoding model also fail to transfer to any of the other models (last row) which shows the difference in the model architecture and the learned feature importance. Lastly, we see that the attack from the (non-regularized) re-upload encoding models is not as effective on the Fourier net ($entry[1,2]$) as in the reversed case ($entry[2,1]$), which opens the question if we can see a change on this matter when we introduce regularization to our re-upload model.

\subsection{Impact of regularization}
As the re-upload encoding architectures as described in Appendix \ref{app-architectures} make use of trainable encodings, we can regularize the input weights as developed by \cite{berberich2023training} and discussed in Section \ref{sec-theo}. Thus we can define variants of the re-upload encoding model named ReUp($\lambda$) with respective regularization rates $\lambda \in \{0.0, 0.1, 0.2\}$ ($\lambda =0.0$ corresponds to no regularization). This way, changes in the Lipschitz bound (by applying regularized training) can be related to the outcomes of transfer attacks and the change in robustness.

\begin{figure}[htb] 
\centering
\includegraphics[width=0.49\textwidth]{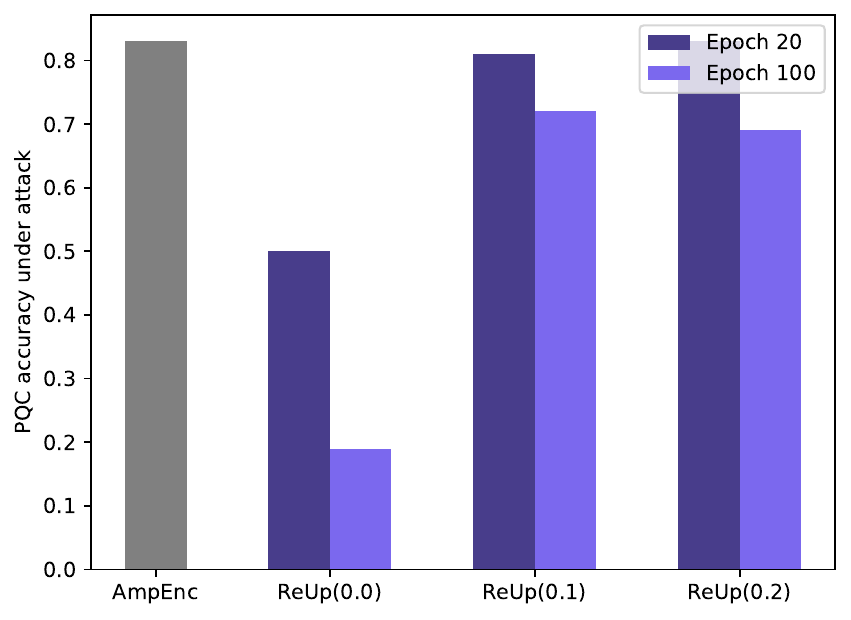}
\caption{Accuracies for transfer attacks from the classical Fourier net to regularized re-upload encoding PQC architectures.}
\label{fig-rff-to-qvc}
\end{figure}

To see this in action, we study transfer attacks between the classical Fourier net and quantum models, where we now include the ReUp($\lambda$) models and check for differences in the success of the transfer attack. We start by using the perturbed images generated by the PGD(0.1) attack constructed for the Fourier net (epoch 20) and feed them into the ReUp($\lambda$) architectures.  
Fig. \ref{fig-rff-to-qvc} shows the resulting accuracy under attack for each of the quantum models with respect to input samples perturbed as described above.

These transfer attacks show that the amplitude encoding model retains its high accuracy, whereas the non-regularized re-upload encoding model performs notably worse. Furthermore, one can see a big difference in the adversarial transfer robustness of the ReUp(0.0) model in relation to the number of trained epochs. Lastly, we can see that regularization helps the quantum models against transfer attacks from classical attacks, where the training epochs have less of an influence.

\begin{figure}[htb]
\centering
\includegraphics[width=0.49\textwidth]{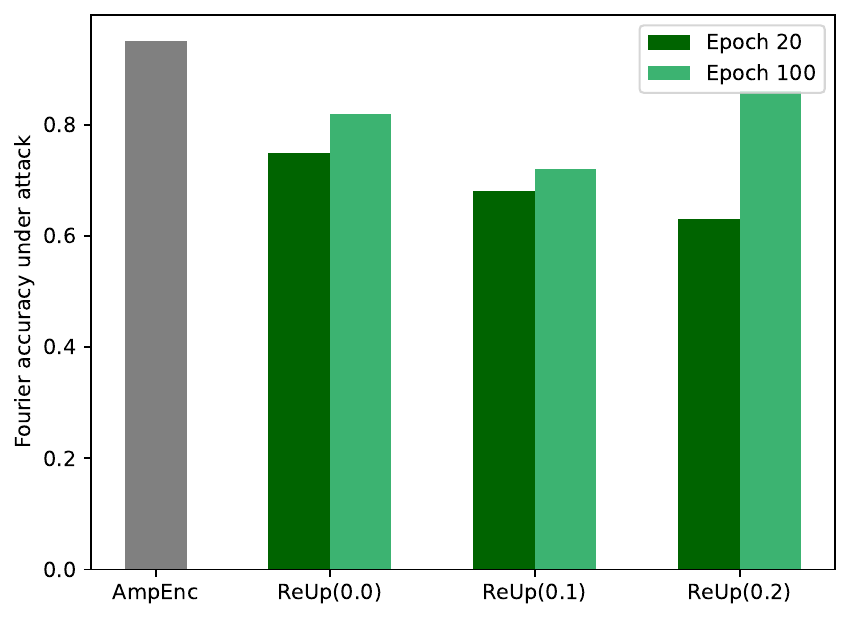}
\caption{Accuracies for transfer attacks from re-upload source models (differing in regularization strength) to the Fourier net target model.}
\label{fig-qvc-to-rff}
\end{figure}

As a logical next step, we investigate the flip side of the experimental coin above: Feeding the perturbed input generated by each of the quantum models into the classical Fourier network trained for 20 epochs. Note that in this case, \textit{we vary the attacks to perturb images (source models), while the target model (Fourier net) stays constant}. 
The green bars in Fig. \ref{fig-qvc-to-rff} visualize the outcome of this switched experiment. The models on the $x$-axis correspond to the source model of the attack, i.e. the model that the perturbations were originally generated for. On the $y$-axis, we can see the accuracy under these transfer attacks for the Fourier net. Again, the first (gray) bar remains high, meaning that the attack from the amplitude encoding model has little impact on the performance of the classical model. 

For the attacks coming from the ReUp($\lambda$) circuits, we notice an increasing success of the attack with higher values of $\lambda$, meaning that the original ReUp($\lambda$) model learns increasingly useful features that are attacked and transfer better over the classical/quantum boundary. It is also apparent that attacks on a model trained for 100 epochs are less successful in fooling the Fourier net, demonstrating that the source model was less potent in focusing on important features.

\begin{figure}[htb] 
\includegraphics[width=0.49\textwidth]{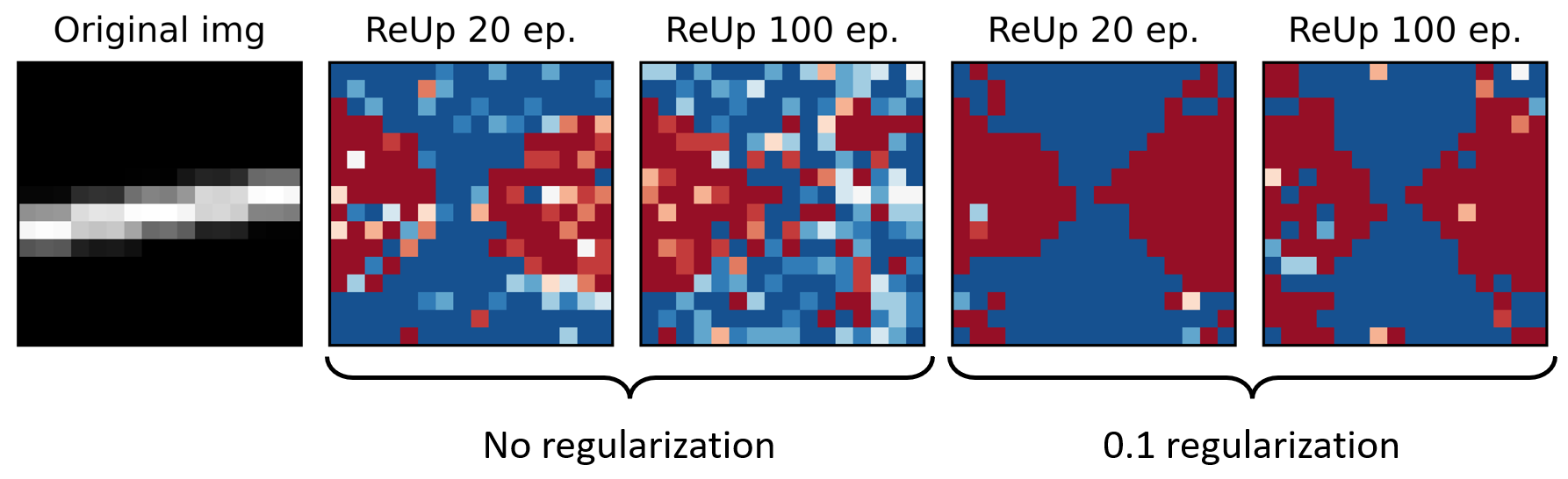}
\caption{Perturbation patterns for PGD($0.1$) attacks on non-regularized versus regularized re-upload encoding models for the sample image shown on the left. The attack pattern for regularized models are less noisy, showing evidence of improved attention to useful features in the underlying model.}
\label{fig-heatmaps}
\end{figure}

Interestingly, a similar pattern emerges when looking at visual modifications $\delta$ for the regularized vs. non-regularized re-upload encoding models. When plotting the raw $\delta$ values of an adversarial input image for each of the models, we can check which pixels are changed the most and in which direction 
(i.e. the value of $\delta_i\in [-\varepsilon,\varepsilon]$). In Fig. \ref{fig-heatmaps}, the resulting perturbation patterns are visualized, where regions in blue correspond to high values ($\delta_i=\varepsilon$) and red areas correspond to low values ($\delta_i=-\varepsilon$). As the original sample belongs to class `$-$', we can see the modifications adding a vertical bar (to flip the predicted label to `$+$') in each case. However, the expected hourglass shape differs in consistency, where the same concept of Fig. \ref{fig-rff-to-qvc} reappears. For non-regularized models, longer training duration has a stark effect on the resulting attack pattern, where the attack on the 100 epoch model looks more chaotic and noisy than the epoch 20 one. If we introduce regularization, the effect becomes less apparent and the overall shape is more defined, meaning that the attack (and thus the underlying model) focuses more on features actually contributing to the target class label.

\subsection{Lipschitz bounds}
To tie the findings of the transfer attack experiments and the perturbation patterns for different regularization strengths together, we return to the introduced measure of Lipschitz bounds. For each model involved in the adversarial attacks earlier, we calculate an approximation of the Lipschitz bound, where we use methods developed by Fazlyab et al.\cite{fazlyab2019efficient} for our classical models and an adapted version of the bound calculated by Berberich et al. \cite{berberich2023training} for the PQC models. As the derived Lipschitz bounds are derived for $\ell_2$ norm constraints, we note that we can use the conservative $\ell_\infty$ conversion multiplying by $\sqrt{D}$, however for the simple comparison between models as we do in the following, this is not needed. Future work may engage in finding tighter bounds and the relations between different model architectures. Also, as discussed by Szegedy et al. \cite{szegedy2013intriguing}, a high Lipschitz bound does not imply the existence of adversarial attack possibilities, but a low bound does give some certificate of robustness. Given the derivation in Appendix \ref{app-lip}, the resulting Lipschitz bounds are stated in Table \ref{table-lipbounds}. 

\begin{table}[h!]
\caption{Approximated Lipschitz bounds for selected models. Regularization of re-upload models has a direct effect on the resulting Lipschitz bound, pushing it towards (or slightly below) the bounds derived for the classical models.}
\renewcommand*{\arraystretch}{1.5}
\centering
\begin{tabular}{| l || c | }
\hline
\multicolumn{1}{|c||}{\textbf{Model}} & \textbf{Lipschitz bound} \\
\hline\hline
ReUp(0.0), epoch 20  & 73.11 \\
ReUp(0.0), epoch 100 & 106.33 \\
\hline
ReUp(0.1), epoch 20  & 18.18 \\
ReUp(0.1), epoch 100 & 19.77 \\
\hline
ReUp(0.2), epoch 20  & 13.50 \\
ReUp(0.2), epoch 100 & 11.83 \\
\hline
Fourier, epoch 20 & 25.38 \\
ConvNet, epoch 20 & 20.24 \\
\hline
\end{tabular}
\medskip

\label{table-lipbounds}
\end{table}

First, the re-upload encoding model shows a relatively large bound, even more so for the increased training time. Second, regularization notably makes a difference, not only in the practical attacks, but also in theory, where $L$ drops to values slightly below the range of those obtained by the analysis of classical models. Lastly, we can see that the additional training time for the re-upload models has less of an influence on $L$ if regularization is introduced during training. These results fit into the picture outlined in the transfer attacks and perturbation patterns above. In the following section, we summarize these findings and put them into place at the current research frontier on QAML.

%% file: 06-summary-outlook.tex
\section{Summary and Outlook}
In this work, a series of experiments has been conducted, which allow for a comparative analysis of adversarial robustness for both quantum and classical machine learning classification networks. The findings of these experiments are summarized below, where we split the results thematically into single-model analysis and inter-model (i.e. transfer) analysis.

\subsection{Adversarial attacks and training epochs}
The main observations of the attack success rate and the perturbation pattern coherence are as follows:

\begin{enumerate}
    \item During training, the features that a re-upload model with trainable encoding attends to change dramatically, which increases Lipschitz bounds on the theoretical side and adversarial vulnerability and perturbation pattern noise on the practical side. For models without trainable encodings, the training time has less/ no influence on this matter, but the resulting model is also less adaptive \cite{berberich2023training}.
    \item If we introduce regularization to the re-upload models, the training time has less impact on robustness as the encoding weights are kept small during training and do not overly emphasize single input features.
    \item The attack perturbation patterns give useful insight into the features that have most influence on the classification decision of a model as these features are perturbed the most. As we repeat the attacks for models of the same architecture with different initializations, this is very similar to popular feature attribution approaches such as SmoothGrad \cite{smilkov2017smoothgrad}.
    \item To this end, the constructed dataset offers useful capabilities into the semantic categorization of useful versus noisy feature perturbations.
\end{enumerate}

\subsection{Transfer attacks}
While single-model attacks offer some insight into robustness and feature attribution of the respective models, the focus of this paper lies on the comparison of robustness for both classical and quantum machine learning models. In this area, the following insights surfaced during the experiments:

\begin{enumerate}
    \item The suspected quantum supremacy in robustness has to be taken with a grain of salt: Many factors influence the success rate of a transfer attack between classical and quantum models. Most importantly, we can see that not all quantum models are robust to classical transfer attacks, which makes regularization an important topic for future research.
    \item Fourier networks, as introduced in this paper, can be seen as ``middle ground'' between classical and quantum networks; attacks from this model architecture transfer well into both quantum and classical domains. Investigating this model class further can bring important insight into attributes of the re-upload encoding model that are otherwise hard to examine.
    \item Amplitude encoding shows greatly different behavior in our experiments than the other models, where transfer attacks are not successful in either direction. While the perturbation pattern seems more noisy for the given attacks, the training time/ regularization has little influence on this measure as the parameters in the model have no input dependency/ influence on input features.
    \item Regularized re-upload models are more successful in transferring attacks to the classical Fourier network. This can be seen in the corresponding Lipschitz bounds, but also in the attack success rate and the attribution to more robust features in the input data.
\end{enumerate}

\subsection{Future work}
While this paper sheds some light on the usefulness of theoretical Lipschitz bounds and practical transfer attacks across the quantum-classical boundary for a comparative robustness analysis paired with perturbation patterns for finding a model's attribution to features, there are still open fields to investigate. First, the calculated Lipschitz bounds for classical and quantum models can be made tighter by considering local bounds given by the absolute modification $\varepsilon$ in each input feature and lower/ upper bounds on the resulting values in the non-linear layers. Also, one can consider $\ell_\infty$ bound certificates to go into more detail of the underlying attacks. On the practical side, quantum classification models are still limited to relatively low-dimensional input data, therefore it would be of research interest to validate the experiments done here on larger-scale input. As a reference, we show in Appendix \ref{app-mnist}, that the re-upload encoding architecture as shown in this paper can be easily adapted to other input datasets such as MNIST. While much work has been dedicated towards re-upload encodings, their Fourier representation, their Lipschitz bounds, regularization and classical approximations, relatively less work is commited to thoroughly finding bounds, classical representations and their implications for comparison of amplitude encoding circuits, which by design should have relatively low Lipschitz bounds resulting from their simple linear structure and direct input encoding into the amplitudes of the circuit qubits. For NISQ devices that are expected to depend on efficient encoding routines such as re-upload encoding, it is nevertheless very important to develop strategies to obtain robustness and generalization guarantees before being able to surpass classical models in practice.

%% file: 07-code.tex
\section{Code availability} \label{sec-code}
The dataset (including train/validation split and data generation script) used in the paper, as well as all models, pretrained weights, and Lipschitz calculations will be made available upon publication.

%% file: 99-appendix.tex
\appendices
\section{Model architectures and training specifications} \label{app-architectures}

\subsection{Quantum model architectures}
\begin{figure}[htb] 
\centering
\includegraphics[width=0.49\textwidth]{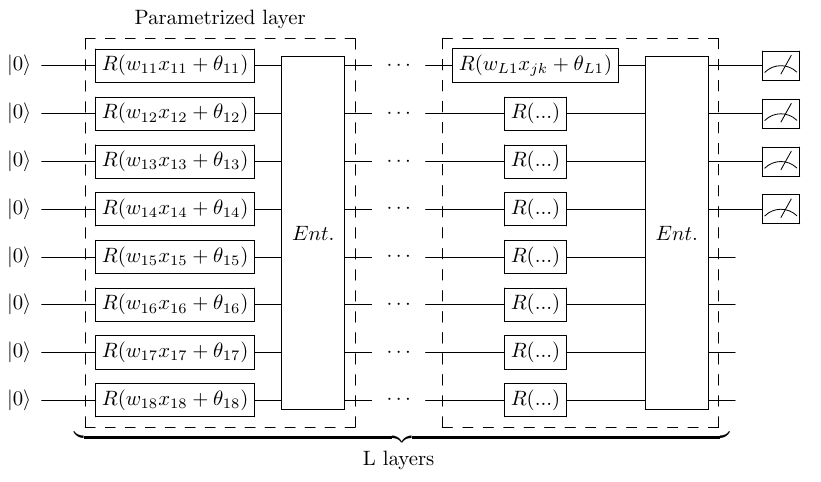}
\caption{Re-upload Encoding circuit used for the experiments in the main text. $L$ \textit{StronglyEntanglingLayers} are used encode the input and optimize over variational parameters, followed by measurements of the first four qubits.}
\label{fig-reup-circ}
\end{figure}

The focus of the paper lies on the re-upload encoding architecture, proposed by P{\'{e}}rez-Salinas et al. \cite{PerezSalinas2020datareuploading}, where the features of the input vector are grouped into 3-tuples and sequentially inserted as rotational angles of (general) rotational gates. A similar architecture is used by Berberich et al. \cite{berberich2023training} (in their training of Lipschitz-regularized quantum models) with the difference that they use an inner product of encoding weights and feature vectors as input to every rotational gate which becomes computationally infeasible for large input dimensions.
Figure \ref{fig-reup-circ} shows a schematic overview over our circuit architecture (based on the \textit{StronglyEntanglingLayers} subroutine), where $w_{lq}\in\mathbb{R}^{3}$ correspond to the encoding weights (for $l\in\{1,\cdots,L\}, q\in\{1,\cdots,Q\}$) to be multiplied with the input features. Each $x_{jk}\in\mathbb{R}^{3}$ therefore corresponds to a 3-tuple of features of the current (linearized) input sample. The expression inside the rotational gates, $w_{lq}x_{jk}$ is then the Hadamard (i.e. element-wise) product of the two vectors with an additional bias vector $\theta_{lq}\in\mathbb{R}^{3}$. In the following, we again shift from the indices used here to denote the position in the circuit to a single index specifying the element of a vector.

Mathematically, given some input 3-tuple $x$ and weights $w,\theta\in\mathbb{R}^{3}$, this general rotation can be written as a product of unitaries 
\begin{equation}
R(wx+\theta)=U_{w,\theta}(x) =\prod_{j=1}^{3} e^{-i(w_j x_j+\theta_j)H_j}
\label{eq-decomp-rot}
\end{equation}
where the typical gate decomposition of general rotational gates (also used in the Pennylane module\footnote{\url{https://docs.pennylane.ai/en/stable/code/api/pennylane.Rot.html}}) uses Hamiltonians $H_1=\frac{1}{2}\sigma_Z$, $H_2=\frac{1}{2}\sigma_Y$, $H_3=\frac{1}{2}\sigma_Z$.

\begin{figure}[htb] 
\centering
\includegraphics[width=0.5\textwidth]{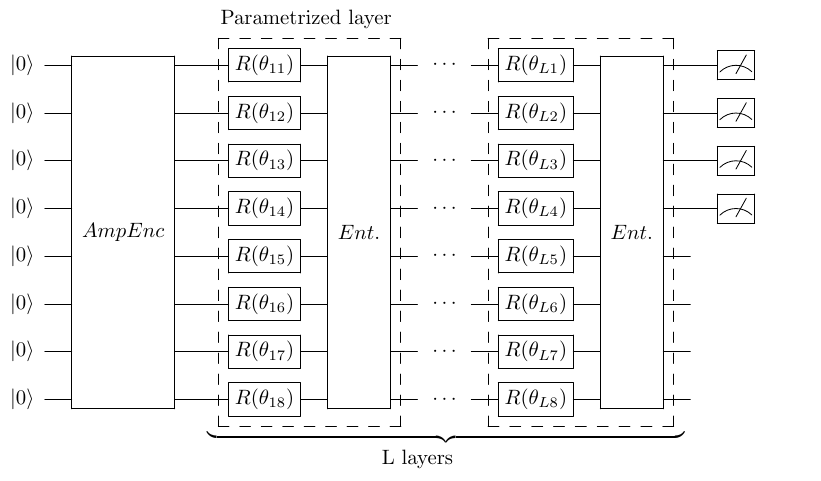}
\caption{Amplitude encoding circuit, where the encoding routine is depicted as a black-box unitary, followed by $L$ parametrized \textit{StronglyEntanglingLayers} and a final measurement of the first four qubits.}
\label{fig-ampenc-circ}
\end{figure}

A similar architecture is used for the amplitude encoding model, where we only use the additive parameters $\theta\in\mathbb{R}^{3}$ inside the general rotational gates as the input is already embedded into the amplitudes of the qubits (shown as a black-box unitary in Fig. \ref{fig-ampenc-circ}). 

For both amplitude and re-upload encoding, we use $L=32$ parametrized \textit{StronglyEntanglingLayers} for $Q=8$ qubits. As model output, we use a softmax distribution from the expectation values of the first four qubits corresponding to the four target classes. For the re-upload encoding model, this means that each feature of a $D=256$-dimensional input sample is shown to the model three times, resulting in a three times re-upload encoding.

\subsection{Classical model architectures}
The classical ConvNet used in the experiments consists of a single convolutional layer (6 output channels, $(5\times5)$-filters), followed by ReLU activation and MaxPool layers. A linear layer maps the obtained latent states to an output vector of dimension four. This simple architecture is easy to analyse while still obtaining sufficiently good performance on the used dataset (this was also verified on MNIST). The model contains a total of $1.024$ parameters, making it comparable to the other model architetures in the experiments.

\begin{figure}[htb] 
\centering
\includegraphics[width=0.47\textwidth]{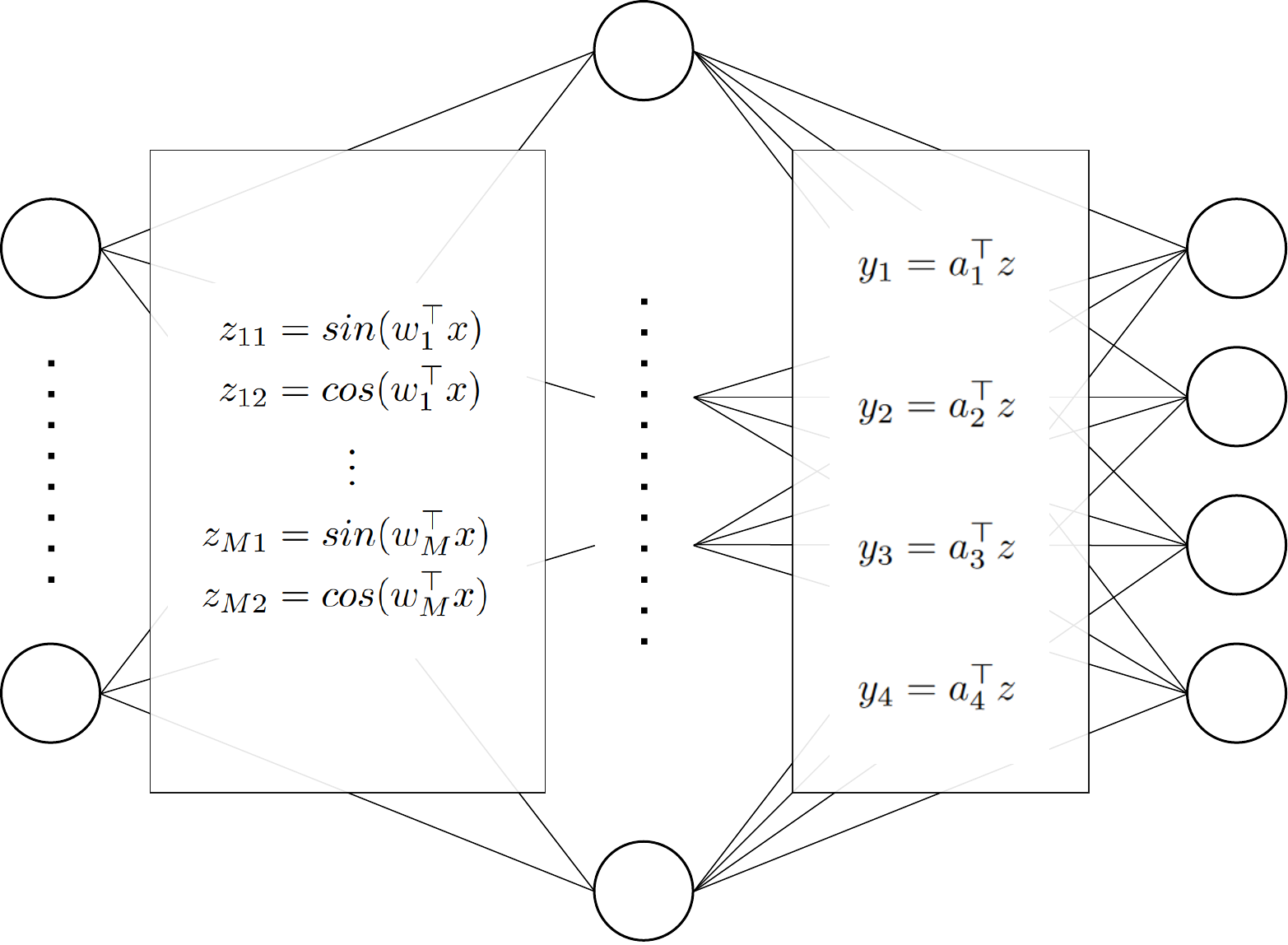}
\caption{Fourier network modeling a classical approximation of re-upload encoding circuits with trainable encodings. A single hidden layer with sine and cosine activation functions connects the $256$-dimensional input to the four-dimensional output.}
\label{fig-fourier-net}
\end{figure}

The Fourier net depicted in Fig. \ref{fig-fourier-net} is built from two linear layers: one embedding layer with $64$ hidden neurons, alternately applying sine and cosine activations and an output layer for the four-dimensional logits of the model. As described in the picture, the weights for two subsequent hidden nodes are shared, therefore the model effectively uses a latent size of $32$ neurons. The transform of the second layer, as denoted in the image $y_i=a_{i}^{\top}z$ contains the trainable coefficients $a_i=(a_{w_1},b_{w_1},\cdots,a_{w_{64}},b_{w_{64}})$ for the summands in the truncated classical Fourier series from \eqref{eq-class-fourier}. The output function therefore directly applies \eqref{eq-class-fourier} to the input samples, with the encoding weights in the first layer and the Fourier coefficients in the output layer.

The code for the models described here can be found in the repository linked in section \ref{sec-code}, alongside the training, optimization and validation routines discussed in the following.

\subsection{Model training}
\begin{figure}[htb] 
\centering
\includegraphics[width=0.49\textwidth]{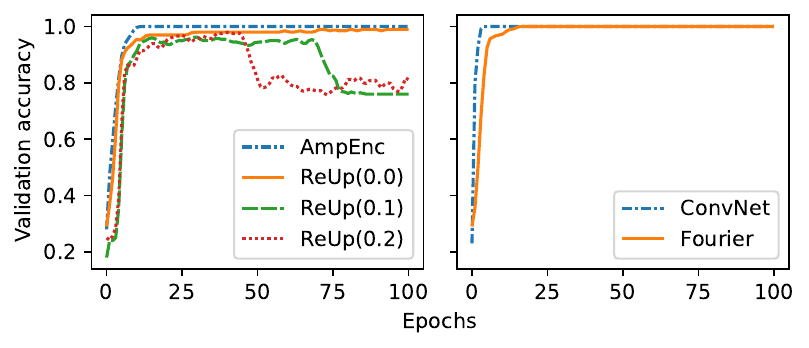}
\caption{Validation accuracy (as we do not fine-tune the training hyperparameters, this can be used to verify the model performance) for the quantum (left) and classical (right) models. For all models, the validation accuracy is $\ge 99\%$ after 10 epochs. The regularized models show a drop in performance due to the regularization term of the loss taking over during training, depending on the parameter $\lambda$.}
\label{fig-val-acc}
\end{figure}

All models are trained using the Adam optimizer routine (0.001 learning rate, batch size 50) in Pytorch; for the quantum models, this means that we first convert the circuits into a Pytorch model using Pennylane's \textit{TorchLayer}\footnote{\url{https://docs.pennylane.ai/en/stable/code/api/pennylane.qnn.TorchLayer.html}} class. After each epoch, the respective model's loss values and accuracy on the training and validation set are recorded. This procedure is repeated 5 times with different initialization seeds to avoid a single ``lucky run''. The averaged model performances are plotted in Fig. \ref{fig-val-acc}. As the re-upload models use encoding weights, we can initialize them (and the biases) with small values to circumvent the Barren plateau problem \cite{mcclean2018barren, grant2019initialization}, as the whole circuit resembles an Identity operation for the first few training steps. Equivalently, we can initialize the variational parameters of the amplitude encoding with small random values to have the same effect.

\section{Derivation of Lipschitz bounds for quantum models} \label{app-lip}
For the key steps in deriving an upper bound on the Lipschitz constant for re-upload encoding models, we refer to the Appendix of Berberich et al. \cite{berberich2023training}. The main difference between their architecture and ours is our use of Hadamard products for the three rotational angles of each general rotational gate and the measurement of a subset of qubits in contrast to a tensor observable over all qubits in their paper (making only a two-class parity measurement possible). The first difference can be adjusted if we consider the decomposition in \eqref{eq-decomp-rot} in its product form, the second difference disappears if we look at each expectation value in our model output $[\langle\sigma_Z^{1}\rangle, \langle\sigma_Z^{2}\rangle, \langle\sigma_Z^{3}\rangle, \langle\sigma_Z^{4}\rangle]$ seperately (as the observables are commuting), which again takes the same form as in \cite{berberich2023training}. Accordingly, the Lipschitz bounds for the re-upload encoding models in this paper are calculated as in \eqref{eq-lip-quant}.

\section{Derivation of Lipschitz bounds for classical models} \label{app-lip-class}
As pointed out in Subsection \ref{subsec-advattacks-lip}, we consider the LipSDP \cite{fazlyab2019efficient} framework for tight upper bounds on the Lipschitz constant of the Fourier net, where the used activation functions (sine, cosine) are slope-restricted by $\alpha=-1$ and $\beta=1$. For the resulting network architecture $f(x)=W^1 \phi\left(W^0 x+b^0\right)+b^1$, where $W^0$ is the encoding weight matrix and $W^1$ is the coefficient weight matrix containing the $a_i$ terms in Fig. \ref{fig-fourier-net}, the resulting LipSDP constrained optimization problem can be found in Thm. 1 of \cite{fazlyab2019efficient}.

For the ConvNet, we use the representation of 2D-convolutions as large Toeplitz block matrices (e.g. described in Araujo et al. \cite{araujo2021lipschitz}) to obtain a simple feed-forward structure as for the Fourier net. As Pytorch implements the 2D-convolution operator as discrete cross-correlations\footnote{\url{https://pytorch.org/docs/stable/generated/torch.nn.Conv2d.html}}, the conversion is straightforward to realize. Unfortunately, the resulting architecture is too computationally expensive for the LipSDP method, however we can still check for the spectral norm of the resulting weight matrices and obtain an upper bound for the Lipschitz bound as detailed in Szegedy et al. \cite{szegedy2013intriguing}, although the bound might not be as accurate as in \cite{fazlyab2019efficient}.

\section{Results for MNIST classification} \label{app-mnist}
All classical and quantum models were also trained on a four-class MNIST classification task, achieving near-perfect training and testing accuracies. The MNIST dataset can be loaded into the models referenced in Section \ref{sec-code} in the same way as our dataset. The experiments in the paper were also successfully conducted on MNIST, however the perturbation patterns are less interpretable due to the nature of the dataset where all regions of the image bear semantic meaning.

%% file: bibliography.bib
@article{sarma2019machine,
    author = {Das Sarma, Sankar and Deng, Dong-Ling and Duan, Lu-Ming},
    title = "{Machine learning meets quantum physics}",
    journal = {Physics Today},
    volume = {72},
    number = {3},
    pages = {48-54},
    year = {2019},
    month = {03},
    issn = {0031-9228},
    doi = {10.1063/PT.3.4164},
    url = {https://doi.org/10.1063/PT.3.4164}
}

@article{schuld2021effect,
  title={Effect of data encoding on the expressive power of variational quantum-machine-learning models},
  author={Schuld, Maria and Sweke, Ryan and Meyer, Johannes Jakob},
  journal={Physical Review A},
  volume={103},
  number={3},
  pages={032430},
  year={2021},
  publisher={APS}
}

@article{schuld2021supervised,
  title={Supervised quantum machine learning models are kernel methods},
  author={Schuld, Maria},
  journal={arXiv preprint arXiv:2101.11020},
  year={2021}
}

@article{vidal2020input,
author={Gil Vidal, Francisco Javier and Theis, Dirk Oliver},   
title={Input Redundancy for Parameterized Quantum Circuits},      
journal={Frontiers in Physics},      
volume={8},           
year={2020},      
url={https://www.frontiersin.org/articles/10.3389/fphy.2020.00297},       
doi={10.3389/fphy.2020.00297},      
issn={2296-424X}  
}

@article{sweke2023potential,
  title={Potential and limitations of random Fourier features for dequantizing quantum machine learning},
  author={Sweke, Ryan and Recio, Erik and Jerbi, Sofiene and Gil-Fuster, Elies and Fuller, Bryce and Eisert, Jens and Meyer, Johannes Jakob},
  journal={arXiv preprint arXiv:2309.11647},
  year={2023}
}

@article{ren2022experimental,
  title={Experimental quantum adversarial learning with programmable superconducting qubits},
  author={Ren, Wenhui and Li, Weikang and Xu, Shibo and Wang, Ke and Jiang, Wenjie and Jin, Feitong and Zhu, Xuhao and Chen, Jiachen and Song, Zixuan and Zhang, Pengfei and others},
  journal={Nature Computational Science},
  volume={2},
  number={11},
  pages={711--717},
  year={2022},
  publisher={Nature Publishing Group US New York}
}

@article{PerezSalinas2020datareuploading,
  doi = {10.22331/q-2020-02-06-226},
  url = {https://doi.org/10.22331/q-2020-02-06-226},
  title = {Data re-uploading for a universal quantum classifier},
  author = {P{\'{e}}rez-Salinas, Adri{\'{a}}n and Cervera-Lierta, Alba and Gil-Fuster, Elies and Latorre, Jos{\'{e}} I.},
  journal = {{Quantum}},
  issn = {2521-327X},
  publisher = {{Verein zur F{\"{o}}rderung des Open Access Publizierens in den Quantenwissenschaften}},
  volume = {4},
  pages = {226},
  month = feb,
  year = {2020}
}

@article{Shin2023exponential,
   title={Exponential data encoding for quantum supervised learning},
   volume={107},
   ISSN={2469-9934},
   url={http://dx.doi.org/10.1103/PhysRevA.107.012422},
   DOI={10.1103/physreva.107.012422},
   number={1},
   journal={Physical Review A},
   publisher={American Physical Society (APS)},
   author={Shin, S. and Teo, Y. S. and Jeong, H.},
   year={2023},
   month=jan }

@article{lu2020quantum,
  title={Quantum adversarial machine learning},
  author={Lu, Sirui and Duan, Lu-Ming and Deng, Dong-Ling},
  journal={Physical Review Research},
  volume={2},
  number={3},
  pages={033212},
  year={2020},
  publisher={APS}
}

@article{west2023benchmarking,
  title={Benchmarking adversarially robust quantum machine learning at scale},
  author={West, Maxwell T and Erfani, Sarah M and Leckie, Christopher and Sevior, Martin and Hollenberg, Lloyd CL and Usman, Muhammad},
  journal={Physical Review Research},
  volume={5},
  number={2},
  pages={023186},
  year={2023},
  publisher={APS}
}

@inproceedings{he2016deep,
  title={Deep residual learning for image recognition},
  author={He, Kaiming and Zhang, Xiangyu and Ren, Shaoqing and Sun, Jian},
  booktitle={Proceedings of the IEEE conference on computer vision and pattern recognition},
  pages={770--778},
  year={2016}
}

@article{szegedy2013intriguing,
  title={Intriguing properties of neural networks},
  author={Szegedy, Christian and Zaremba, Wojciech and Sutskever, Ilya and Bruna, Joan and Erhan, Dumitru and Goodfellow, Ian and Fergus, Rob},
  journal={arXiv preprint arXiv:1312.6199},
  year={2013}
}

@inproceedings{papernot2017practical,
  title={Practical black-box attacks against machine learning},
  author={Papernot, Nicolas and McDaniel, Patrick and Goodfellow, Ian and Jha, Somesh and Celik, Z Berkay and Swami, Ananthram},
  booktitle={Proceedings of the 2017 ACM on Asia conference on computer and communications security},
  pages={506--519},
  year={2017}
}

@inproceedings{demontis2019adversarial,
  title={Why do adversarial attacks transfer? explaining transferability of evasion and poisoning attacks},
  author={Demontis, Ambra and Melis, Marco and Pintor, Maura and Jagielski, Matthew and Biggio, Battista and Oprea, Alina and Nita-Rotaru, Cristina and Roli, Fabio},
  booktitle={28th USENIX security symposium (USENIX security 19)},
  pages={321--338},
  year={2019}
}

@article{ilyas2019adversarial,
  title={Adversarial examples are not bugs, they are features},
  author={Ilyas, Andrew and Santurkar, Shibani and Tsipras, Dimitris and Engstrom, Logan and Tran, Brandon and Madry, Aleksander},
  journal={Advances in neural information processing systems},
  volume={32},
  year={2019}
}

@article{madry2017towards,
  title={Towards deep learning models resistant to adversarial attacks},
  author={Madry, Aleksander and Makelov, Aleksandar and Schmidt, Ludwig and Tsipras, Dimitris and Vladu, Adrian},
  journal={arXiv preprint arXiv:1706.06083},
  year={2017}
}

@article{west2023towards,
  title={Towards quantum enhanced adversarial robustness in machine learning},
  author={West, Maxwell T and Tsang, Shu-Lok and Low, Jia S and Hill, Charles D and Leckie, Christopher and Hollenberg, Lloyd CL and Erfani, Sarah M and Usman, Muhammad},
  journal={Nature Machine Intelligence},
  volume={5},
  number={6},
  pages={581--589},
  year={2023},
  publisher={Nature Publishing Group UK London}
}

@article{landman2022classically,
  title={Classically Approximating Variational Quantum Machine Learning with Random Fourier Features},
  author={Landman, Jonas and Thabet, Slimane and Dalyac, Constantin and Mhiri, Hela and Kashefi, Elham},
  journal={arXiv preprint arXiv:2210.13200},
  year={2022}
}

@article{schreiber2023classical,
  title={Classical surrogates for quantum learning models},
  author={Schreiber, Franz J and Eisert, Jens and Meyer, Johannes Jakob},
  journal={Physical Review Letters},
  volume={131},
  number={10},
  pages={100803},
  year={2023},
  publisher={APS}
}

@article{caro2021encoding,
  title={Encoding-dependent generalization bounds for parametrized quantum circuits},
  author={Caro, Matthias C and Gil-Fuster, Elies and Meyer, Johannes Jakob and Eisert, Jens and Sweke, Ryan},
  journal={Quantum},
  volume={5},
  pages={582},
  year={2021},
  publisher={Verein zur F{\"o}rderung des Open Access Publizierens in den Quantenwissenschaften}
}

@article{virmaux2018lipschitz,
  title={Lipschitz regularity of deep neural networks: analysis and efficient estimation},
  author={Virmaux, Aladin and Scaman, Kevin},
  journal={Advances in Neural Information Processing Systems},
  volume={31},
  year={2018}
}

@inproceedings{araujo2021lipschitz,
  title={On lipschitz regularization of convolutional layers using toeplitz matrix theory},
  author={Araujo, Alexandre and Negrevergne, Benjamin and Chevaleyre, Yann and Atif, Jamal},
  booktitle={Proceedings of the AAAI Conference on Artificial Intelligence},
  volume={35},
  number={8},
  pages={6661--6669},
  year={2021}
}

@article{fazlyab2019efficient,
  title={Efficient and accurate estimation of lipschitz constants for deep neural networks},
  author={Fazlyab, Mahyar and Robey, Alexander and Hassani, Hamed and Morari, Manfred and Pappas, George},
  journal={Advances in Neural Information Processing Systems},
  volume={32},
  year={2019}
}

@article{pauli2021training,
  title={Training robust neural networks using Lipschitz bounds},
  author={Pauli, Patricia and Koch, Anne and Berberich, Julian and Kohler, Paul and Allg{\"o}wer, Frank},
  journal={IEEE Control Systems Letters},
  volume={6},
  pages={121--126},
  year={2021},
  publisher={IEEE}
}

@article{berberich2024robustness,
  title={Robustness of quantum algorithms against coherent control errors},
  author={Berberich, Julian and Fink, Daniel and Holm, Christian},
  journal={Physical Review A},
  volume={109},
  number={1},
  pages={012417},
  year={2024},
  publisher={APS}
}

@article{berberich2023training,
  title={Training robust and generalizable quantum models},
  author={Berberich, Julian and Fink, Daniel and Pranji{\'c}, Daniel and Tutschku, Christian and Holm, Christian},
  journal={arXiv preprint arXiv:2311.11871},
  year={2023}
}

@article{gong2022enhancing,
  title={Enhancing Quantum Adversarial Robustness by Randomized Encodings},
  author={Gong, Weiyuan and Yuan, Dong and Li, Weikang and Deng, Dong-Ling},
  journal={arXiv preprint arXiv:2212.02531},
  year={2022}
}

@inproceedings{gallant1988there,
  title={There exists a neural network that does not make avoidable mistakes},
  author={Gallant},
  booktitle={IEEE 1988 International Conference on Neural Networks},
  pages={657--664},
  year={1988},
  organization={IEEE}
}

@article{smilkov2017smoothgrad,
  title={Smoothgrad: removing noise by adding noise},
  author={Smilkov, Daniel and Thorat, Nikhil and Kim, Been and Vi{\'e}gas, Fernanda and Wattenberg, Martin},
  journal={arXiv preprint arXiv:1706.03825},
  year={2017}
}

@article{paszke2019pytorch,
  title={Pytorch: An imperative style, high-performance deep learning library},
  author={Paszke, Adam and Gross, Sam and Massa, Francisco and Lerer, Adam and Bradbury, James and Chanan, Gregory and Killeen, Trevor and Lin, Zeming and Gimelshein, Natalia and Antiga, Luca and others},
  journal={Advances in neural information processing systems},
  volume={32},
  year={2019}
}

@article{bergholm2018pennylane,
  title={Pennylane: Automatic differentiation of hybrid quantum-classical computations},
  author={Bergholm, Ville and Izaac, Josh and Schuld, Maria and Gogolin, Christian and Ahmed, Shahnawaz and Ajith, Vishnu and Alam, M Sohaib and Alonso-Linaje, Guillermo and AkashNarayanan, B and Asadi, Ali and others},
  journal={arXiv preprint arXiv:1811.04968},
  year={2018}
}

@article{lecun2010mnist,
  title={MNIST handwritten digit database},
  author={LeCun, Yann and Cortes, Corinna and Burges, CJ},
  journal={ATT Labs [Online]. Available: http://yann.lecun.com/exdb/mnist},
  volume={2},
  year={2010}
}

@article{mcclean2018barren,
  title={Barren plateaus in quantum neural network training landscapes},
  author={McClean, Jarrod R and Boixo, Sergio and Smelyanskiy, Vadim N and Babbush, Ryan and Neven, Hartmut},
  journal={Nature communications},
  volume={9},
  number={1},
  pages={4812},
  year={2018},
  publisher={Nature Publishing Group UK London}
}

@article{grant2019initialization,
  title={An initialization strategy for addressing barren plateaus in parametrized quantum circuits},
  author={Grant, Edward and Wossnig, Leonard and Ostaszewski, Mateusz and Benedetti, Marcello},
  journal={Quantum},
  volume={3},
  pages={214},
  year={2019},
  publisher={Verein zur F{\"o}rderung des Open Access Publizierens in den Quantenwissenschaften}
}
